\pdfoutput=1
\documentclass[11pt]{article}

\usepackage[final]{acl}

\usepackage{times}
\usepackage{latexsym}

\usepackage[T1]{fontenc}
\usepackage[utf8]{inputenc}

\usepackage{microtype}

\usepackage{inconsolata}

\usepackage{microtype}
\usepackage{hyperref}
\usepackage{url}
\usepackage{booktabs}
\usepackage{inconsolata}
\usepackage{graphicx}

\usepackage{url}
\usepackage{tabularx}
\usepackage{graphicx} 
\usepackage{subcaption} 
\usepackage{multirow}
\usepackage{ulem}
\usepackage{amsmath}
\usepackage{algorithm}
\usepackage{algorithmic}
\usepackage{tabularx} % For adjustable width tables
\usepackage{wrapfig}
\usepackage{enumitem}
\usepackage{xcolor}
\usepackage{placeins}
\usepackage{listings}

\usepackage{tablefootnote}
\usepackage{authblk}
\usepackage[utf8]{inputenc}
\usepackage[most]{tcolorbox}
\usepackage{enumitem}

\title{LLM-Coordination: Evaluating and Analyzing Multi-agent Coordination Abilities in Large Language Models}

\author{
    \textbf{Saaket Agashe, Yue Fan, Anthony Reyna, Xin Eric Wang} \\
    University of California, Santa Cruz \\
    \texttt{\{saagashe, yfan71, ancreyna, xwang366\}@ucsc.edu}
}

\begin{document}
\maketitle
\begin{abstract}
Large Language Models (LLMs) have demonstrated emergent common-sense reasoning and Theory of Mind (ToM) capabilities, making them promising candidates for developing coordination agents. This study introduces the LLM-Coordination Benchmark, a novel benchmark for analyzing LLMs in the context of \textbf{Pure Coordination} Settings, where agents must cooperate to maximize gains. Our benchmark evaluates LLMs through two distinct tasks. The first is \emph{Agentic Coordination}, where LLMs act as proactive participants in four pure coordination games. The second is \emph{Coordination Question Answering (CoordQA)}, which tests LLMs on 198 multiple-choice questions across these games to evaluate three key abilities: Environment Comprehension, ToM Reasoning, and Joint Planning. Results from Agentic Coordination experiments reveal that LLM-Agents excel in multi-agent coordination settings where decision-making primarily relies on environmental variables but face challenges in scenarios requiring active consideration of partners' beliefs and intentions. The CoordQA experiments further highlight significant room for improvement in LLMs' Theory of Mind reasoning and joint planning capabilities. \emph{Zero-Shot Coordination (ZSC)} experiments in the Agentic Coordination setting demonstrate that LLM agents, unlike RL methods, exhibit robustness to unseen partners. These findings indicate the potential of LLMs as Agents in pure coordination setups and underscore areas for improvement. Code Available at \url{https://github.com/eric-ai-lab/llm_coordination}.  
\end{abstract}

\begin{figure*}[h!]
    \centering
    \includegraphics[width=\textwidth]{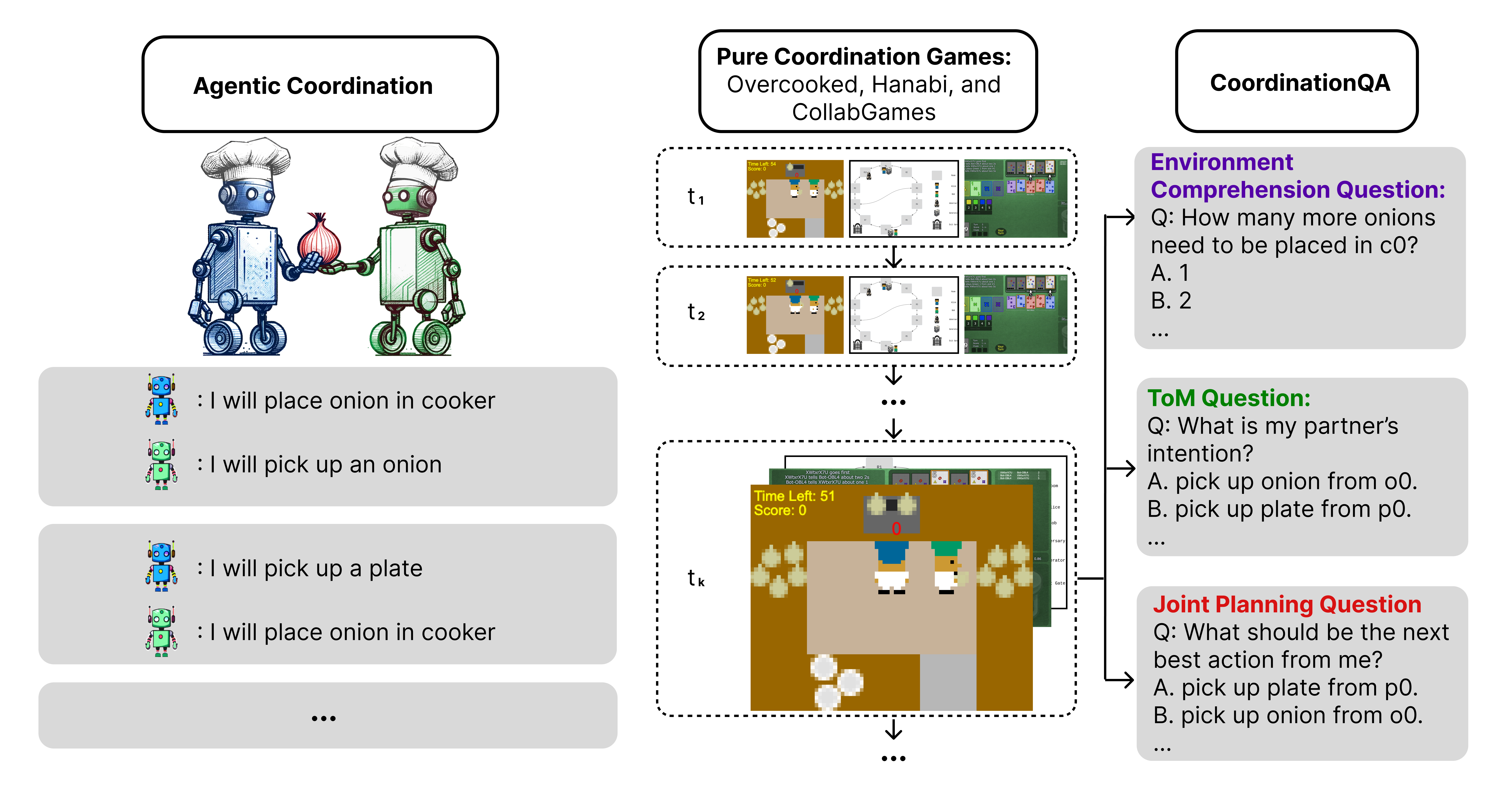}
    \caption{\textbf{The LLM Coordination Benchmark} consists of two tasks: \textit{Agentic Coordination} to study the holistic abilities of LLMs in multi-turn pure coordination games, and \textit{Coordination QA} to perform a fine-grained analysis of the Environment Comprehension, Theory of Mind Reasoning, and Joint Planning abilities of LLMs in the context of pure coordination scenarios.
    }
    \label{fig:llm_co_benchmark}
\end{figure*}

% \saaket{simplify sentence structure}
\section{Introduction}
% \saaket{Emphasize pure coordination settings - The games are by nature pure coordination rather than Evaluation of fundamental coordination abilities of LLMs not of Agent Frameworks. Allowing more straight forward comparison to RL-methods for Multiagent Coordination}
In a wide range of activities, from daily tasks such as cooking to critical operations like rescue efforts, cooperation without mixed intentions is essential. These scenarios are examples of Pure Coordination Games, where all involved parties benefit from choosing strategies that are perfectly aligned, avoiding any conflict of interest. These games require agents to reason about their environment and plan while considering the beliefs and intentions of their partners. Recently, Large Language Models (LLMs) have demonstrated emergent planning abilities in both physical and virtual settings \citep{raman2022planningCorrectivePrompting, wang2023voyager, wu2023spring}, impressive reasoning capabilities \citep{wei2022chain}, and the hints of a Theory of Mind \citep{kosinski2023theory} making them promising candidates for developing coordination agents. Previous works have explored the use of LLMs for developing collaborative agents, yet the requisite conditions, strengths, and limitations of LLMs in coordination games remain unclear. In this study, we intend to bridge the gap by performing a comprehensive evaluation and analysis of the multi-agent coordination abilities of LLMs.

% \saaket{No more reason vs. act formulation. More like unit test / abl study - component analysis}
Therefore, we introduce the \textbf{LLM-Coordination Benchmark} featuring two task settings for pure coordination games: 1. Agentic Coordination and 2. CoordinationQA. In Agentic Coordination, LLMs are scaffolded with components that allow them to act within actual game environments, providing a holistic evaluation of the competencies of LLMs to act as coordination agents. In CoordinationQA, LLMs have to answer a curated set of questions about edge-case scenarios drawn from coordination games where agents need to actively cooperate with their partners.  
The benchmark includes four collaborative games, providing a comprehensive analysis platform.
Unlike studies on multi-LLM frameworks~\cite{hong2023metagptmetaprogrammingmultiagent, qian2024chatdevcommunicativeagentssoftware, li2023camelcommunicativeagentsmind}, which focus on orchestrating multiple LLMs to solve tasks, our benchmark assesses the innate ability of individual LLMs to understand and act within pure coordination scenarios where cooperation is essential.

Our experiments in the Agentic Coordination setting reveal that Large Language Models are competent at understanding the game objectives, generating coherent reasoning for their next actions, and coordinating with partners across all coordination games. They exhibit these behaviors without any training, fine-tuning, or few-shot examples. A comparative analysis reveals that LLM agents match or outperform RL baselines in games where optimal decision-making can be done by observing environment variables and positions (e.g., Overcooked). However, they struggle in settings where agents need to actively consider their partner’s beliefs and intentions (e.g., Hanabi). We also observe that LLM agents are capable of collaborating with new partners, unlike self-play MARL methods~\citep{onTheUtility, hanabi} that fail to adapt to unseen agents.  
 
For a more nuanced analysis of the coordination abilities of LLMs, we create the CoordinationQA Suite. This suite is designed to dissect the capabilities of LLMs in single-turn reasoning within coordination games, focusing on three key areas: Joint Planning, Theory of Mind (ToM), and Environment Comprehension. Joint Planning (JP) evaluates LLMs' planning abilities for optimal coordination, ToM questions probe their understanding of partner agents' intentions and needs, and Environment Comprehension (EC) assesses their inference of environment details, rules and objectives.
First, Our findings on CoordinationQA show a marked performance gap between GPT-4-turbo and other LLMs across three question types. Secondly, LLMs are most proficient in Environment Comprehension, indicating they understand the rules and environment states well. However, they face significant challenges in Theory of Mind Reasoning, with difficulty inferring others' intentions and needs. This issue worsens in Joint Planning, where most LLMs underperform, some even worse than random choices. These results highlight LLMs' limited reliability and effectiveness as coordination partners. Correlation analysis between LLMs' performance on CoordinationQA and their performance on agentic coordination setting further highlights their strengths in environmental reasoning but exposes significant weaknesses in Theory of Mind inference and Joint Planning capabilities.

In summary, our contributions are threefold: 

\begin{enumerate}[leftmargin=*]
    \item We introduce the LLM-Coordination Benchmark for evaluating and analyzing LLMs in Pure Coordination Games, covering multi-turn Agentic Coordination and single-turn Coordination QA tasks.
    \item We perform a holistic evaluation of LLM agents in Self-play and Cross-play settings, offering a detailed comparison with RL baselines and showcasing their potential as Coordination Agents.
    \item We investigate Environment Comprehension, Theory of Mind Reasoning, and Joint Planning as essential components of LLMs' overall coordination capabilities, highlighting their critical importance in pure coordination setups.
\end{enumerate}

\section{Related Work}

\paragraph{Multi-agent Coordination.}
Pure Coordination games in game theory are scenarios where agents share the payoffs, and cooperation is the optimal strategy. Benchmarks like the Multiparticle Environment~\cite{mpe}, Overcooked-AI~\cite{onTheUtility}, and the Hanabi Challenge~\cite{hanabi} evaluate multi-agent coordination. \citet{onTheUtility} highlighted the importance of human data for effective Human-AI collaboration. Subsequent Overcooked-AI research focuses on aligning self-play-trained agents with humans using techniques such as self-play with past checkpoints~\citep{fcp}, population entropy objectives~\citep{mep}, graph-theoretic objectives~\citep{cole}, policy ensembles~\citep{pecan}, and integrating human biases~\citep{hsp}. In the Hanabi Challenge, efforts aim to learn grounded policies over arbitrary conventions~\cite{hu2021otherplay, hu2021offbelief}. While most solutions enhance RL methods for coordination, we propose that LLMs offer an alternative due to their emergent reasoning and theory-of-mind-like abilities, avoiding arbitrary joint interactions.

\paragraph{Planning and Reasoning with Large Language Models.}
LLMs have shown remarkable natural language reasoning abilities~\citep{openai2023gpt4, chatgpt, vicuna2023}, achieving state-of-the-art results in verbal reasoning tasks. Augmented with components like memory and tools, LLMs can interact with external environments, solving long-horizon tasks and playing complex games~\citep{wu2023spring, wang2023voyager, codeaspolicies2022, song2022llm}. Guided by Cognitive Architectures for Language Agents~\citep{sumers2023cognitive} as a design principle for agent design, we experiment with advanced reasoning strategies such as ReAct~\cite{yao2023reactsynergizingreasoningacting}, Self-Verification~\cite{selfverif}, and Self-Consistency~\cite{wang2023selfconsistencyimproveschainthought} to enhance LLM reasoning. These strategies establish strong baseline performance for our Language Agent implementations.

\paragraph{Multi-agent LLMs.}
Recent studies have explored LLMs in multi-agent cooperation settings. \citet{zhang2023buildingEmbodiedModular} developed a modular agent framework for spatial rearrangement tasks. \citet{zhang2023proagent} introduced an architecture enabling LLMs to play Overcooked-AI. \citet{shi2023cooperationflyexploringlanguage} demonstrated positive zero-shot coordination in Avalon using code-driven reasoning. \citet{li2023metaagents} showed emergent collaborative abilities of LLMs in simulations, while \citet{Li_2023} investigated theory-of-mind inference using explicit belief representations. \citet{xu2024exploringlargelanguagemodels} perform an analysis of LLMs in communication games \citet{xu2023magicinvestigationlargelanguage} analyzed LLM cognitive abilities through games, highlighting benefits of probabilistic modeling. In contrast, our research rigorously evaluates LLM agents' coordination abilities in established pure coordination games, where coordination is essential. Our setting only includes scenarios where agents must fully cooperate with each other with no competitive incentives. We also conduct a fine-grained, component-level analysis to understand the intricacies of LLMs' coordination capabilities.

% Embodied environments usually set up in household environments have also been recently used to study multi-agent coordination \citep{puig2021watchandhelp, CordialSync, TwoBody, gan2021threedworld}.
% \saaket{Some of these works are from after our first submission of the paper for review. We might be able to add some of those works in discussion and show explicitly the differences. add works mentioned by R2 to this part}

% \saaket{general rationale for selection of pure-coordination games as testbeds}
\section{LLM-Coordination Benchmark}

\subsection{Multi-turn Agentic Coordination}

In the Multi-turn Agentic Coordination task, LLMs participate in end-to-end pure coordination games as agents, where the best strategy for all participating agents is to cooperate. We only consider pure coordination scenarios with no competitive incentives. LLMs under test are plugged into coordination frameworks with memory and the ability to act in complete games. These \textbf{LLM agents} can then be partnered with any policies or agents to complete the games.

Our LLM-Coordination benchmark includes 4 pure coordination games: Hanabi Challenge \citep{hanabi}, Overcooked-AI \citep{onTheUtility}, and Collab Capture and Collab Escape (inspired by the Pursuit-Evasion problem). These games were carefully selected for their ability to isolate and highlight specific coordination challenges, providing controlled environments that allow the analysis of pure coordination scenarios without too much emphasis on other reasoning challenges. While LLMs are versatile and capable of addressing a wide range of tasks, these well-studied settings offer established benchmarks, clear metrics, and reproducible scenarios that are particularly suited for examining coordination abilities of participating agents. 
\label{sec:pure_coordination_games}
% Our LLM-Coordination benchmark includes 4 pure coordination games, Hanabi Challenge \citep{hanabi}, Overcooked-AI \citep{onTheUtility}, Collab Capture and Collab Escape.

\label{sec:pure_coordination_games}
\paragraph{Hanabi Challenge.} In Hanabi \citep{hanabi}, players aim to assemble five sequences of cards in ascending order (1 through 5), each sequence dedicated to a different color: purple, red, blue, yellow, and green. 
A unique aspect of the game is that the players can only view their partner's cards, not their own. This requires players to work collaboratively, utilizing reveal tokens to provide hints about the cards in their partner's hand. These hints can be about either the color or the rank of the cards. For instance, using a single reveal token, a player can indicate all cards of a certain rank in their partner's hand. Once a player has an idea about which card they have, they can choose to play the card on the stack. If the card is correct, they get a point. Otherwise, players lose a collective life token. The loss of all 3 life tokens leads to the end of the game. Hanabi serves as an exemplary Pure Coordination game, necessitating player cooperation to achieve optimal outcomes. Success in Hanabi hinges on the ability to understand partners' perspectives, navigate decisions based on incomplete information, and engage in implicit communication, making it an excellent testing ground for coordination among agents. 

\paragraph{Overcooked-AI.} In the Overcooked-AI environment \citep{onTheUtility}, two agents—Alice (Blue) and Bob (Green)—collaborate to cook and deliver onion soups. This environment includes a variety of layouts, each with its own arrangement and quantity of onion dispensers, plate dispensers, cookers, delivery zones, and countertops. To prepare a dish, agents are required to insert three onions into a cooker, initiating a cooking process that lasts 20 time steps. Upon completion, the soup must be plated and delivered to complete the task. Each layout presents unique challenges, emphasizing the need for agents to comprehend their surroundings, locate necessary resources, and synchronize their actions with their teammate for effective collaboration. 

\paragraph{Collab Capture.} Collab Capture involves two agents trying to capture an adversary in a maze of interconnected rooms. The rooms are connected by doors, which can be controlled through access buttons that can be found in other rooms. The agents' task is to capture the adversary in the least amount of time using effective strategies. To evaluate different coordination strategies between agents, we design four scenarios by controlling the states of the doors (open or closed) within the layout shown in figure \ref{fig:collab_map}. These scenarios highlight various coordination challenges, such as trapping the adversary through precise positioning, enabling a teammate by prioritizing door control over direct pursuit, and strategically restricting the adversary’s movement to facilitate capture. (see Appendix \ref{app:collabCapture} for more details.)

\paragraph{Collab Escape.} Collab Escape involves two agents trying to escape an adversary in a maze of interconnected rooms. They need to fix two generators (similar to the game Dead-by-Daylight \citep{DeadByDaylight2016}) located in different rooms to open an exit portal. The adversary tries to catch the agents, and the win condition is any one agent escaping. To evaluate coordination strategies in Collab Escape, we develop two scenarios by varying the initial proximity of agents to the adversary and generators in the layout shown in figure \ref{fig:collab_esc_map}. Depending on their proximity to the adversary/generators, players need to apply strategies such as luring the adversary away from
the partner, choosing to continue fixing the generators while sacrificing for the partner’s safety and manipulating the movement of the adversary (see Appendix \ref{app:collabEscape} for more details.)

\subsection{Single-turn Coordination QA}

The agentic coordination task paints a holistic picture of the abilities of LLMs as agents. To dive deeper into the specific strengths and weaknesses of LLMs, we develop the CoordinationQA Suite. Inspired by the idea of \textbf{Unit Testing} for evaluating AI agents \cite{knott2021evaluatingRobustness}, we manually sampled edge cases from all 4 pure coordination games mentioned in Section \ref{sec:pure_coordination_games}. All of these edge cases necessitate agents to actively understand their current state, think about their partner's intentions, and come up with the best plans for coordination.  We then create a set of three types of questions for each scenario in our CoordinationQA Suite.

\begin{itemize}[leftmargin=*, itemsep=0pt, topsep=0pt]
    \item \textbf{Environment Comprehension (EC)} questions require LLMs to make indirect inferences about some aspect of their environment (See Appendix \ref{app:ec}). The questions cover details of the layouts, implications of current observations, and counts of artifacts.  
    \item \textbf{Theory of Mind Reasoning (ToM)} questions challenge the LLMs to predict the intentions of their partners and probe about the requirements of their partners (See Appendix \ref{app:tom}). 
    \item \textbf{Joint Planning (JP)} questions provide agents with the state/observation and ask them to predict the best next action for effective coordination. This question is essentially the same question that LLMs need to repeatedly solve when they act as agents (See Appendix \ref{app:jp}). 
\end{itemize}

% \begin{figure*}[h]
%     \centering
%     \includegraphics[width=\textwidth]{assets/CAC_j.png}
%     \caption{LLM Agents for agentic coordination. We segment the process into: Memory, which archives the game description and current game state; Grounding, which involves the execution of actions selected by LLMs; and Reasoning, which can optionally include a Theory of Mind (ToM) inference LLM, a verifier LLM, and the primary LLM under analysis.}
%     \label{fig:cac}
% \end{figure*}

All the questions were manually developed and labeled. We filtered out questions and scenarios that showed any ambiguity, leaving only questions that had clear, optimal solutions. We generated a total of N=66 scenarios (25 from Overcooked, 28 from Hanabi, and 13 from the two Collab Games) and created 3 questions per scenario, resulting in 198 unique questions. The right side of Figure \ref{fig:llm_co_benchmark} demonstrates the sampling process for the three types of questions with an example from the game Overcooked. The selected scenario shows the Blue agent about to place their third onion in the cooker, and the green agent needs to figure out what to do next. See Appendix \ref{app:gqc} for examples of questions and the templates used to formulate these questions.

\section{Experimental Setup}
\subsection{Agentic Coordination}

% \subsubsection{Setup}

We perform two types of experiments in agentic coordination: Self-Play and Cross-Play. In self-play settings, the participating agents are of the same type. In Cross-Play experiments, we pair agents with unseen partners, and they need to adapt their behavior to the actions of these new partners.  

\subsubsection{LLM Agents}
To allow LLMs to play multi-turn games, we scaffold them with an agentic framework based on Cognitive Architectures for Language Agents \citet{sumers2023cognitive}. The framework includes three parts: Memory, Reasoning, and Grounding. % (See Figure \ref{fig:cac})

\textbf{Memory} includes (1) Long-Term Memory for storing the Game Description, including the game's rules, conventions, objectives, and action space, (2) Working memory, which consists of a textual description of the current observation, and (3) Episodic Memory which is a list of previous actions selected by the agent. 

\textbf{Reasoning} is where the Large Language Model (LLM) is plugged into the framework. It takes the textual description from the Memory as input and generates the next action based on the context. The LLM reasons about the current state and then selects an action from the list of available actions in natural language. 

\emph{Self-Verification: } For the coordination game Hanabi, there is a low margin for error as any misplays lead to the loss of life tokens, and the loss of all three life tokens subsequently results at the end of the game. We thus supplement the reasoning process in Hanabi with Answer-Verification \cite{selfverif}, where the LLM is re-prompted to confirm that the action it generated is appropriate and does not lead to fatal errors.

\emph{ToM-Reasoning: } We also demonstrate the positive impact of a Theory of Mind Reasoning step prior to generating the next action for Hanabi and CollabEscape, which benefit from this intermediate step. In the ToM reasoning step, the LLM generates an interpretation of their partner's actions or current position before generating the next action to explicitly capture the belief inference process. We do not test with additional ToM reasoning on Overcooked due to significant latency and cost constraints, with marginal benefits. 

Finally, the \textbf{Grounding} process translates the natural language action generated by the reasoning module into game-compatible action(s). The exact implementation of the grounding module depends on the game in question; for example, in Overcooked-AI, the grounding module needs to convert high-level actions like "pick up onion from o0." into sequences of lower-level actions. On the other hand, in games like Hanabi, the Grounding needs to match actions like "Reveal Bob's Red Color Cards" to their lower-level representations. The Grounding process is also responsible for filtering out infeasible actions based on the context of the game (See Appendices \ref{sec:overcooked-appendix}, \ref{sec:hanabi-appendix} for more details.)

There are no prompt or setup differences for LLM Agents based on Cross-play or Self-play. We use the LLMs gpt-4-0125-preview, GPT-3.5-turbo-0125, Mixtral 8x7B, and GPT-4o for agentic evaluation studies.

\begin{table*}[h]
    \centering
    \setlength{\tabcolsep}{3pt} 
    \small
    \resizebox{\textwidth}{!}{%
    \begin{tabular}{lllllll}
        \toprule
        % & \multicolumn{5}{c}{\textbf{Overcooked Layouts}} \\
        % \cmidrule(lr){2-6}
        \textbf{Agent} & CR & AA & Ring & FC & CC \\
        \midrule
        Self-Play (PPO)& $198.8 \pm 4.06$ & $167.2 \pm 3.63$ & $\mathbf{190.8 \pm 4.25}$ & $151.9 \pm 3.28$ & $122.3 \pm 3.80$ \\
        PBT& $\mathbf{216.9 \pm 1.31}$ & $190.1 \pm 8.64$ & $173.8 \pm 18.27$ & $169.5 \pm 10.09$ & $140.1 \pm 13.86$ \\
        \midrule
        GPT-3.5-turbo & $33.3 \pm 10.88$ & $46.6 \pm 10.88$ & $40.0 \pm 0.00$ & $66.6 \pm 14.40$ & $53.3 \pm 5.44$ \\
        Mixtral8x7B & $46.6 \pm 14.40$ & $200.0 \pm 9.42$ & $113.3 \pm 5.44$ & $46.6 \pm 14.40$ & $100.0 \pm 9.42$ \\
        GPT-4o & $160 \pm 0.00$ & $166.66 \pm 5.44$ & $66.66 \pm 21.77$ & $120.0 \pm 9.42$ & $\mathbf{160.0 \pm 0.00}$ \\
        GPT-4-turbo & $173.3 \pm 6.67$ & $\mathbf{260.0 \pm 11.55}$ & $140.0 \pm 0.00$ & $\mathbf{180.0 \pm 11.55}$ & $\mathbf{160.0 \pm 0.00}$ \\
        \bottomrule
    \end{tabular}
    }
    \caption{Performance comparison across Multi-Agent Reinforcement Learning (MARL) and LLM-agent methods. Scores indicate the best performance in each category. The GPT-4-turbo Agent demonstrates superior coordination in 3 out of 5 scenarios, underscoring advanced reasoning capabilities in coordination tasks. The five layouts are CR: Cramped Room, AA: Asymmetric Advantages, Ring: Coordination Ring, FC: Forced Coordination, and CC: Counter Circuit. For visualization and details of these layouts, see appendix \ref{sec:overcooked-appendix}}
    \label{tab:ai-ai-comparison}
\end{table*}

\begin{table*}[!t]
\centering
\small
\begin{tabular}{lcccc}
\toprule
& \multicolumn{2}{c}{\textbf{Collab Escape}} & 
\multicolumn{2}{c}{\textbf{Collab Capture}} \\
\cmidrule(lr){2-3} \cmidrule(lr){4-5}

\textbf{Agent} & Capture Rate & Avg. Turns & Escape Rate & Avg. Turns \\ 

\midrule
GPT-4-turbo    & 0.83 & 4.60 & 1.00 & 3.5 \\
-- (w/out ToM Reasoning)   & 0.50 & 2.00 & 1.00 & 4.75 \\
GPT-4o    & 0.67 & 4.00 & 1.00 & 7.17 \\
GPT-3.5-turbo  & 0.33 & 2.50 & 0.67 & 8.38  \\
Mixtral-8x7b   & 0.50 & 7.67 & 0.92 & 7.55 \\

\midrule
Greedy Baseline& 0.00 & N.A. &  0.50 & 6.00 \\
\bottomrule
\end{tabular}
\caption{Comparison of different LLM Agents on CollabCapture and CollabEscape. In CollabEscape, two agents work together to escape from an adversary. In CollabCapture, two agents coordinate to capture an adversary. The reported results are run across 3 trials each for various layout configurations (Detailed in Appendix \ref{app:collab}). The table also demonstrates the impact of the explicit ToM reasoning step in both game setups.}
\label{tab:cc}
\end{table*}

\subsubsection{MARL Agents}

\textbf{Self-play MARL Baselines:} For Overcooked we use Proximal Policy Optimization \citep{schulman2017proximal} and Population-Based Training \citep{jaderberg2017population} as baselines for comparison. These baselines were established by \citet{onTheUtility}. 

For the Hanabi challenge, we use Bayesian Action Decoder (BAD) \citep{hanabi}, Simplified Action Decoder (SAD) \citep{hu2021simplified}, and Off-Belief Learning \citep{hu2021offbelief} as MARL baselines for Hanabi. All three baselines achieve near-perfect performance in Self-play. 

\textbf{Cross-play MARL Baselines:} For Overcooked, we use a Behavior Cloning model trained on human data \cite{onTheUtility} and a Proximal Policy Optimization (PPO) agent trained with the Human Behavior Cloning agent \cite{onTheUtility} as baselines for comparison. We also report Hidden-Utility Self-play (HSP) \cite{hsp} as a baseline. We use human proxies based on behavior cloning as unseen partners.

For Hanabi, we use the Simplified Action Decoder (SAD), which is trained through self-play as a baseline. We pair our agents with Off-Belief Learning \citep{hu2021offbelief}, which was trained to generate grounded policies and adapt to unseen partner agents.

\paragraph{Metrics.} We measure the total score achieved by agents in Overcooked, where each delivery provides 20 points to both agents. In the case of Hanabi, the metric is the total number of cards that have been correctly arranged by the players. For CollabEscape and CollabCapture, we report the success rate of escape or capture across multiple trials and the average turns to capture or escape.

% \textbf{LLM Agents: } There are no setting differences for LLM Agents based on Cross-play or Self-play. We keep the same procedural prompts and abilities for both settings. For Overcooked, we report our results using ReAct \cite{yao2023reactsynergizingreasoningacting} prompting with GPT-4-turbo, GPT-3.5-turbo, and Mixtral8x7b models for Overcooked. In Hanabi, we report results for LLMs with ReAct, ReAct + Self-Verification \cite{selfverif} and ToMReAct + Self-verification. We define ToMReAct (ToM inference+Reason+Act) as a variant of ReAct where we use an explicit Theory of Mind reasoning step to interpret partner clues and requirements before reasoning about the next action.

\subsection{CoordinationQA}

We assess the performance of 5 Families of Large Language Models (LLMs) \cite{jiang2023mistral, jiang2024mixtral, touvron2023llama, vicuna2023, openai2023gpt4} across three dimensions: Environment Comprehension (EC), Theory of Mind Reasoning (ToM), and Joint Planning (JP). For each category, LLMs respond to multiple-choice questions (MCQs), with their responses evaluated against ground-truth answers through fuzzy string matching. To account for the variability in LLM responses, we conduct three trials per model. All models being tested are shown the same prompts. We also report a Random baseline. 

% \saaket{add gpt-4o and gpt-4o-mini to the tables}
% \saaket{Discuss citations in main method
% Add more LLMs
% [x] Add methods to ZSC section
% Justify the choice of baselines in the response and the paper 
% [x] Add proAgent}
\section{Discussion}

% \begin{table*}[h!]
%     \centering
%     \small
%     \begin{tabular}{lcccc}
%     \toprule
%     & \multicolumn{2}{c}{\textbf{Collab Capture}} & 
%     \multicolumn{2}{c}{\textbf{Collab Escape}} \\
%     \cmidrule(lr){2-3} \cmidrule(lr){4-5}
%     \textbf{Method} & Capture Rate & Turns to Capture & Escape Rate & Turns to Escape \\ 
    
%     \midrule
%     ReAct\textsubscript{GPT-4-turbo}    & 1.00 & 3.99 & 1.00 &24.67 \\
%     ReAct\textsubscript{GPT-3.5-turbo}  & 0.50 & 8.49 & 0.00 & N.A.  \\
%     ReAct\textsubscript{Mixtral-8x7b}   & 0.75 & 3.88 & 0.00 & N.A. \\
%     Greedy Baseline& 0.50 & 6.00 & 0.00 & N.A. \\
%     \bottomrule
%     \end{tabular}
%     \caption{Comparison of different LLMs on CollabCapture and CollabEscape with the ReAct framework. The ReAct\textsubscript{GPT-4-turbo} agents achieve 100\% success rate on both games. However, other LLMs fail on CollabEscape games, which also take a long time for GPT-4-turbo to complete.}
%     \label{table:model_combined}
% \end{table*}

\paragraph{Zero-shot LLM Agents match or surpass trained RL methods in Environment-focused Coordination Problems.}

\label{sec:agentic_exps}
We observed that LLM agents (w. GPT-4-turbo) outperform or match the overall performance of RL methods across all layouts of Overcooked-AI. 
Table \ref{tab:ai-ai-comparison} presents the numerical scores attained by different agents when paired with a partner agent of the same type. 
This implies that LLM agents match RL agents that have been explicitly trained through Self-play without any game-specific training or fine-tuning. However, it is important to note that LLM agents are significantly slower and larger than RL models, making them unsuitable for real-time use at present. We also see positive results on the CollabCapture and CollabEscape games, with most LLMs being able to complete both challenges (see Table \ref{tab:cc}).

\label{sec:cra}

% Specifically, GPT-4-turbo exhibits an average latency of $8.36 \pm 1.79$ seconds with chain-of-through and $1.02 \pm 0.09$ seconds without it in our configuration. Other models we tested—GPT-3.5-turbo and Mixtral8x7b—are faster but do not meet the performance of the RL baselines.
\begin{table}[t]

\centering
\setlength{\tabcolsep}{2pt} 
\begin{tabular}{lr}
\toprule
\textbf{Agent} & \textbf{Score} \\
\midrule
Bayesian Action Decoder & \(23.92 \pm 0.01\) \\
Simplified Action Decoder & \(24.01 \pm 0.01\) \\
Off-Belief Learning & \(24.10 \pm 0.01\) \\
\midrule
GPT-4-turbo & \(\mathbf{13.33} \pm 0.88\) \\
- (w.o ToM Reasoning) & \(10.33 \pm 0.88\) \\ 
- (w.o ToM Reasoning \& Verif.) & \(4.33 \pm 0.88\) \\
GPT-4o & \(8.33 \pm 1.20\) \\ 
GPT-3.5-turbo & \(1.33 \pm 0.72\) \\
Mixtral-8x7b & \(0.33 \pm 0.27\) \\
\bottomrule
\end{tabular}
\caption{Agentic performance comparison on Hanabi Challenge. RL methods are very strong and obtain near-perfect scores. The best GPT-4-turbo-based LLM Agent is much weaker compared to RL baselines. Removing the ToM reasoning and Verification steps from the LLM agent leads to further performance degradation}
\label{tab:hanabi_full}
\end{table}

\begin{figure}[t]
    \centering
    \includegraphics[width=\columnwidth]{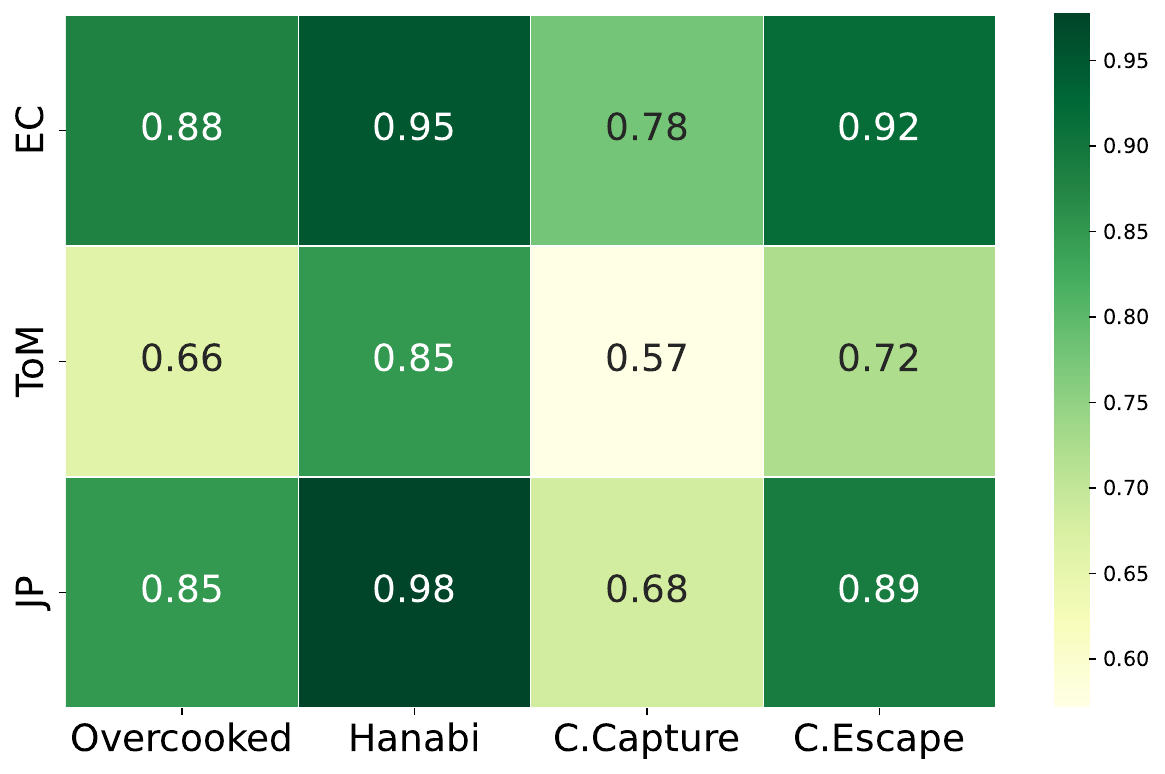}
    \caption{Correlation of LLM Agent performance in Agentic Coordination setup on all four games \textit{vs.} performance on the CoordinationQA benchmark.}
    \label{fig:corr}
\end{figure}

\begin{figure*}[t]
    \centering
    \includegraphics[width=\textwidth]{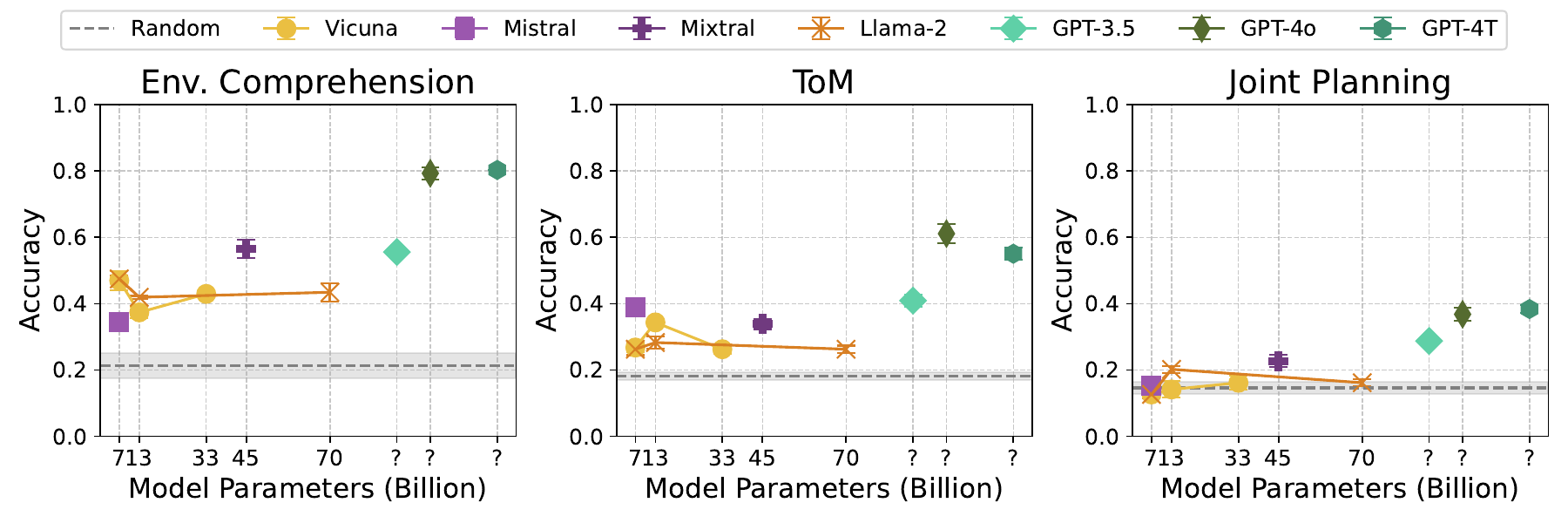}
    \caption{The performance of different LLMs on \textit{CoordinationQA}, which provides a fine-grained analysis of LLMs' Environment Comprehension, Theory of Mind Reasoning, and Joint Planning abilities within pure coordination scenarios.}
    \label{fig:stt}
\end{figure*}

\paragraph{LLM agents struggle at effective planning when advanced Theory of Mind reasoning is required.~}

In Hanabi Challenge, LLM agents seem to struggle compared to RL methods (see Table ~\ref{tab:hanabi_full}). GPT-4-turbo performs reasonably well, while other LLMs can barely complete the games. We attribute this failure to two factors. First, there is little room for errors in Hanabi. Any misplay leads to the loss of a life token. Second, Hanabi requires more complex Theory of Mind Reasoning compared to the Overcooked-AI environment. Each action requires agents to actively consider their partner’s beliefs, intentions, and how they would react to implicit communication. In contrast, Overcooked is fully observable, and its action space consists of actions like \textit{pick up an onion from onion\_dispenser\_0} and \textit{place onion in cooker\_0}. Under most scenarios and layouts, LLMs only need to consider the next best steps based on the state of the environment.

We use correlation study to provide additional validation to these findings. We calculate the Pearson Correlation Coefficient ($r$) of the performance of the four LLMs (GPT-4-turbo, GPT-4o, GPT-3.5-turbo, and Mixtral 8x7b) on Agentic Coordination setup (Average score per game) \textit{vs.} the score on CoordinationQA task. Figure \ref{fig:corr} shows a high correlation between Environment Comprehension capabilities and Success at Overcooked, but a more moderate correlation between Theory of Mind Reasoning capabilities and Success at Overcooked. Conversely, in Hanabi, high success is strongly correlated with both Environment Comprehension and Theory of Mind Reasoning Abilities. In CollabEscape we see a higher correlation with ToM reasoning abilities compared to CollabCapture. This correlation study also connects and establishes a positive alignment between LLM-agent performance on the multi-turn agentic task and the single-turn CoordinationQA.

\begin{table*}[ht]
    \centering
    \small
    \setlength{\tabcolsep}{5pt} 
    \resizebox{\textwidth}{!}{%
    \begin{tabular}{lcccccc}
        \toprule
        \textbf{Method} & CR & AA & Ring & FC & CC \\
        \midrule
        BC & $103.5$ \texttt{|} $110.0$ & $136.5$ \texttt{|}  $137.5$ & $59.0$ \texttt{|}  $70.0$ & $20.5$ \texttt{|}  $31.0$ & $38.0$ \texttt{|}  $44.0$ \\
        PPO$_{BC}$ & $156.4$ \texttt{|} $\mathbf{163.9}$ & $72.6$ \texttt{|} $178.8$ & $126.4$ \texttt{|} $129.8$ & $58.9$ \texttt{|} $76.9$ & $69.5$ \texttt{|}  $57.6$ \\

        HSP & - & $\mathbf{300.3}$ \texttt{|} $\mathbf{217.1}$ & $160.0$ \texttt{|} $\mathbf{160.6}$ & - & $107.4$ \texttt{|} $\mathbf{106.6}$ \\
        
        GPT-4-turbo\footnotemark  & $\mathbf{160.0}$ \texttt{|} $160.0$& $180.0$ \texttt{|} $200.0$ & $\mathbf{160.0}$ \texttt{|} $140.0$ & $\mathbf{120.0}$ \texttt{|} $\mathbf{80.0}$ & $\mathbf{140.0}$ \texttt{|} $100.0$\\
        \bottomrule
    \end{tabular}
    }
    \caption{Zero-shot coordination results of AI-Human Proxy Gameplay. We compare Behavior Cloning (BC), PPO\_BC, HSP \cite{hsp}, and GPT-4-turbo agent. The LLM agent outperforms the PPO and BC methods and matches the HSP \cite{hsp} baseline in most cases, demonstrating robustness to unseen partner agents. Since the two agents in Overcooked-AI might be tasked with different roles based on their starting locations, we show results playing from either side separated by \texttt{|}.}
    \label{tab:ai-human-proxy-comparison}
\end{table*}

\begin{table*}[h]
\centering
\begin{tabular}{lccc}
\toprule
\textbf{Method} & \textbf{Self-Play} & \textbf{Cross-Play w/ OBL-1} & \textbf{Cross-Play w/ OBL-4} \\
\midrule
SAD   & $\mathbf{23.66} \pm 0.54$    & $11.33 \pm 4.00$   & $8.00 \pm 0.47$    \\
GPT-4-turbo   & $13.66 \pm 0.27$ & $\mathbf{15.00} \pm 2.94$ &  $\mathbf{12.00}\pm 0.94$ \\
\bottomrule
\end{tabular}
\caption{Cross-Play results of RL agent (SAD) and LLM agent (GPT-4-turbo). All agents play three games with different seeds (same seeds across agents). SAD performs really well at self-play but suffers significant performance degradation with new partners OBL-1 and OBL-4. LLM Agents coordinate well with the new, unseen partners.}
\label{tab:hanabi_zsc}
\end{table*}

\paragraph{Auxiliary reasoning strategies like Verification and ToM reasoning help LLMs reason for coordination.~}
Adding an Answer Verification step significantly reduces fatal mistakes (wrong card plays) caused by LLM hallucinations. Without the support of the Verification step, LLM agents bomb (lose all three lives) before the end of the game in every trial. The ToM reasoning step separates the tasks of interpreting partner clues and generating actions, allowing the LLM to better synthesize available information for action planning. Table~\ref{tab:hanabi_full} shows the impact of ablating the verification and ToM reasoning steps from the LLM Agent. The ToM reasoning step is also useful in the CollabEscape (see table \ref{tab:cc}) game, as players need to actively consider what their partner needs and act sacrificially if needed. In CollabCapture, it shows a relatively low benefit since agents can observe the positions of all agents as well as doors on the map and infer the correct action based on this environmental context. 

% MARL agents trained with self-play struggle when paired with unseen partners in common payoff tasks, because they converge to arbitrary policies that only the two partners involved in the self-play training understand \citep{onTheUtility, hanabi}. Since LLM agents haven't been explicitly trained to play these games, they base their outputs on the provided textual observation and commonsense knowledge learned from pre-training, and thus are much more robust to unseen partners.  

% \subsection{Coordination QA}

% \subsection{LLM Agents excel at CollabCapture but can struggle with CollabEscape}
% \label{app:collab}

% Table \ref{table:model_combined} summarizes the performance of the three LLMs GPT-4-turbo, GPT-3.5-turbo and Mixtral-8x7b on CollabGames - Collab Capture and Collab Escape. GPT-4-turbo achieves a 100\% success rate in both games. In general, CollabEscape was a more difficult game to solve than CollabCapture since agents needed to perform sacrificial moves and lure the killer away, requiring more advanced Theory of Mind inference. 

\paragraph{Comparative Results of LLMs in Environment Comprehension, ToM Reasoning, and Joint Planning.~}
\label{sec:coordqa}
In Figure \ref{fig:stt}, we see that most LLMs achieve their best results on the Environment Comprehension question. The best performing LLM GPT-4-turbo gets more than 80\% Environment Comprehension Questions correct. The overall performance across LLMs drops on the more challenging Theory of Mind reasoning questions. Both GPT-4-turbo and GPT-4o do well on the Theory of Mind reasoning questions. The overall accuracy of LLMs on Joint Planning questions is still significantly weak, with even the best LLM scoring less than 40\%, indicating a large room for improvement in LLMs' ability to perform coordination reasoning. Another cause for concern is that open-source LLMs perform abysmally at Joint Planning, with some models performing worse than random.

% \textbf{Connecting CoordinationQA benchmark to multi-turn agentic results}

% As shown in Table \ref{tab:correlation_results}, evaluations on the coordination reasoning benchmark validates that Overcooked relies more on Environment Comprehension abilities, while Hanabi requires both Environment Comprehension and strong Theory of Mind Abilities. Furthermore, it also validates that Joint Planning questions serve as a strong proxy for Overall coordination abilities of LLMs. 

% ProAgent \cite{zhang2023proagent} 

% \saaket{Add convention learning experiments - maybe we can also incorporate these into CoordinationQA instead of full agentic games, isn't that the advantage of coordqa?}
\paragraph{LLM Agents are robust to unseen partners.~}

We use Overcooked-AI and the Hanabi challenge as testbeds to evaluate the performance of LLM agents when paired with unseen agents. This task is popularly known as \textit{Zero Shot Coordination}. In Overcooked-AI, we pair our LLM agents as well as baselines with proxy-human agents. These proxy human agents are behavior cloning agents trained using human data by \citet{overcookedAI}. As shown in Table \ref{tab:ai-human-proxy-comparison}, we discover that LLM agents outperform both Behavior Cloning as well as PPO agents trained with human data. Apart from the Asymmetric Advantages layout, they also match or outperform the Hidden Utility Self-play (HSP) baseline, which is designed to excel at ZSC. 
\footnotetext{For GPT-4-turbo, we run a single trial from either position due to cost and time constraints.}

In Hanabi, we pair our agents with Off-Belief Learning (OBL) agents \citep{hu2021offbelief}. OBL is a MARL strategy that generates grounded clues and actions and is the state-of-the-art method for cross-play in Hanabi. OBL agents provide observation-grounded clues and collaborate well with humans. Therefore, we use them as unseen partners in our experiments.
Table \ref{tab:hanabi_zsc} shows that the GPT-4-turbo agent scores an average of 15 points with the OBL-1 agent compared to their self-play scores of 13.66, indicating no degradation in performance with a new partner. The baseline RL method, Simplified Action Decoder (SAD) \cite{hu2021simplified}, fails critically when paired with unseen OBL agents, even though it excels at self-play (22.00 points) due to self-play training.

\section{Conclusion}
In this study, we evaluated and analyzed the current large language models in the context of pure coordination games. We introduced the LLM-Coordination benchmark with its two tasks: 1. Agentic Coordination and 2. CoordinationQA. These settings allowed us to conduct holistic comparative studies of LLMs as agents and dive deeper into the fine-grained aspects of LLMs as coordination reasoners. We juxtaposed LLM agents with existing Multi-agent Reinforcement Learning agents, discussing the conditions in which LLMs thrive and fail. Finally, we discussed the Theory of Mind Reasoning, Environment Comprehension, and Joint Planning as prerequisites for coordination and evaluated existing LLMs on these components. 

\section{Acknowledgements}
This research project has benefitted from the Microsoft Accelerate Foundation Models Research (AFMR) grant program.

\section{Limitations}

Latency and Compute Requirements: As highlighted in Section \ref{sec:agentic_exps}, effective reasoning for coordination is achievable primarily with larger LLMs like GPT-4-turbo. However, these models are associated with significant latency and require substantial computational resources, making them less suitable for real-time applications where rapid decision-making is crucial. 

Initial Prompt Configuration: Achieving optimal reasoning performance necessitates careful manual configuration of the initial prompts that describe the game (Procedural Memory). While this prompt could be extracted from game manuals or existing resources, it still needs to be formatted and designed with the LLM agent in mind. Furthermore, the results for individual games could be improved by letting the LLM generate more text and engineering the prompt. However, we leave these optimizations to future works focused on performance improvement rather than benchmarking. 

Manual Curation of Edge Cases: The CoordinationQA suite involves manually curating unambiguous edge cases in coordination games to construct the dataset. This can hinder the ability to scale the benchmark to accommodate new scenarios. Yet, it is important to curate these examples for more reliable studies.

\bibliography{custom}

\appendix

\begin{figure*}[t]
    \centering
    \includegraphics[width=\textwidth]{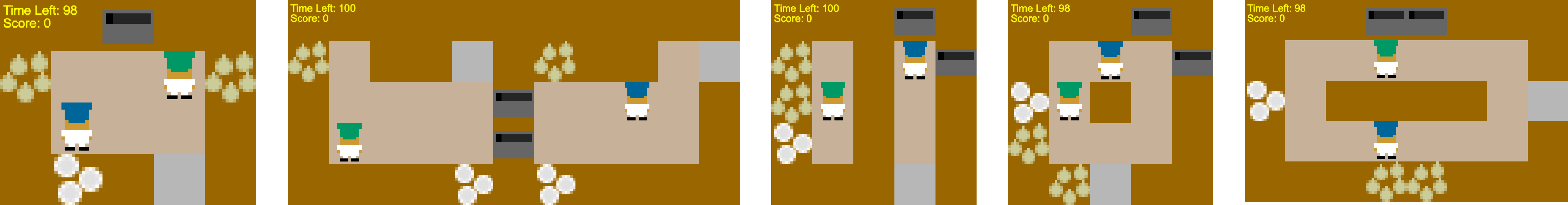}
    \caption{The Overcooked layouts from left to right: Cramped Room (CR), Asymmetric Advantages (AA), Forced Coordination (FC), Coordination Ring (CR), and Counter Circuit (CC).}
    \label{fig:layouts}
\end{figure*}
\section{Overcooked Implementation Details}

In the game Overcooked-AI, two chefs must coordinate their actions to cook and deliver onion soups. Each soup requires three onions as ingredients, which can be found in onion dispensers. The players must add three onions to a cooker to start cooking. Once the three onions are added, the soup starts cooking automatically and requires 20 time steps to complete. Once the soup is cooked, a player needs to pick up a plate, load the soup, and deliver it to the required delivery area. Players must coordinate their actions to effectively deliver soup. Depending on the layouts (see figure \ref{fig:layouts}), players need to adapt to cramped spaces, understand environment layouts for role assignments, pass objects to each other, make way for one another, and find the most effective paths. 
\label{sec:cra}

\label{sec:overcooked-appendix}
\subsection{Game and Layout Description} We use a general game description ${G}$ that explains the rules and objectives of overcooked. Since each layout has a different number of locations, like onion dispensers and cookers, we include a succinct description of each environment ${L_i}$, which includes how many instances of particular facilities there are. For environments that include partitions, we mention which partition each of the agents is situated in and what facilities that agents can access. In addition, we also mentioned the shape of the environment. 

\begin{figure}[H]
    \centering
    \noindent \begin{minipage}{\columnwidth}
\begin{lstlisting}[breaklines=true]
I am {self.player_names[self.player_id]}. I am playing the 
game Overcooked with my partner {self.player_names[self.other_player_id]}. 
{EnvDescriptions[self.layout_name]}
Overcooked has the following rules: {self.rules}. 
We have agreed to follow the following conventions: 
{self.conventions}. I\'ll provide my action history, 
current state, teammate's status, and my possible actions.
Help me select the best action from the list. 
Format your response as: 
Explanation:<Brief explanation for my next action>. 
Action: <action>.
Only select one action. Do not say anything else. Got it?
\end{lstlisting}
\end{minipage}
\vspace{-.5em}
    % \caption{CAPTION}
    \label{fig:appendix-figure1}
\end{figure}

\subsection{State Description}
The State is represented in natural language \( D(S) \) in the working memory, which can be processed by a Large Language Model (LLM). The state \( S \) includes variables that fully represent the necessary details of the layout as well as the players. The information provided in \( D(S) \) is equivalent to what would be accessible to a Reinforcement Learning (RL) agent in the form of state representations. The following information is included in \( D(S) \):

\paragraph{Objects Held by Each Player}
The state description \( D(S) \) begins by detailing the inventories 
 $I_{\alpha_1}$ and $I_{\alpha_2}$ of Alice and Bob, respectively. Each inventory $I_{\alpha_i}$(where \( i \in \{1, 2\} \)) can contain one of the following items: \{"onion", "plate", "cooked soup"\}. This inventory information is translated into natural language and incorporated into \( D(S) \) in the format: ``I am holding $I_{\alpha_1}$. Bob is holding $I_{\alpha_2}$.'' Such information is vital for inferring the likely subsequent actions of the partner agent.

\paragraph{Location of the Agent Controlled by LLM:}
Given the limitations of Large Language Models (LLMs) in interpreting grid-based spatial information, we opt to provide processed location data to the LLM. For each agent \( P_i \) (where \( i \in \{1, 2\} \)), and for each location of interest denoted as \( \text{loc} \), we calculate the distance \( d_{(P_i, \text{loc})} \) as the number of steps required to reach \( \text{loc} \) from \( P_i \) using the shortest available path. The state description \( D(S) \) then includes this processed location information in the format: ``\text{loc} is \( d_{(P_i, \text{loc})} \) units away.'' Here, \( \text{loc} \) can represent various points of interest such as onion dispensers, plate dispensers, cookers, delivery areas, kitchen counters, or shared counters. If a location is either inaccessible or blocked by another agent, this is explicitly stated in \( D(S) \). For example, if a location is blocked by Bob, it would be stated as ``\text{loc} is blocked by Bob.'' To distinguish between the location information relevant to each agent, \( D(S) \) prefixes the respective sections with ``Your location information:'' for the agent controlled by the LLM and ``Bob's location information:'' for the partner agent.

\noindent\paragraph*{Cooker Information}
The state description \( D(S) \) also incorporates information about the cooker, which is central to the gameplay strategy. Specifically, for each cooker \( i \), \( D(S) \) includes the number of onions \( n_{\text{i}} \) currently in the pot. Additionally, \( D(S) \) provides the operational state of the cooker, denoted as \( \text{CookerState}_i \), which can be either "Off" or "On". Lastly, the current condition of the soup in the cooker is represented by \( \text{SoupState}_i \), which can take one of the following values: "Cooking", "Cooked", or "Not Started". Thus, the information for cooker $c_i$ is formatted as: ``\( c_i \) has \( n_{\text{i}} \) onions. \( c_i \) is \( \text{CookerState}_i \). Soup in \( c_i \) is \( \text{SoupState}_i \).'' 

\noindent\paragraph*{Kitchen Counter Information}
The state description \( D(S) \) includes information about kitchen counters, which are primarily used for temporary object storage. Specifically, \( D(S) \) identifies the closest empty kitchen counter \( k_{\text{empty}} \) and the set \( K_{\text{filled}} \) 
of all counters currently holding an object.

\noindent\paragraph*{Shared Counter Information}
Shared counters serve as specialized kitchen counters for object transfer between agents. For each shared counter \( i \), \( D(S) \) includes the status for \( s_i \), as ``\( s_0 \) is empty'' or ``\( s_1 \) contains onion,'' to offer a complete environmental overview. Unlike kitchen counters, where only the closest empty counter is mentioned, all empty shared counters are mentioned. 

% \clearpage

\begin{figure}[H]
    \centering
\noindent 
\begin{minipage}{\columnwidth}
\begin{lstlisting}[]
<Inventory>: I am holding onion. Bob is holding nothing. 

<My Location Information>: o0 is 0 units away. o1 is 1 units away. p0 is 3 units away. c0 is 6 units away blocked by Bob. c1 is 7 units away. d0 is 4 units away. s0 is 1 units away. s1 is 0 units away. s2 is 1 units away. s3 in 2 units away. Closest empty kitchen counter k12 is 1 units away. 

<Bob's Location Information>: o0 is blocked by Alice. o1 is 7 units away. p0 is 3 units away. c0 is 0 units away. c1 is 1 units away. d0 is 4 units away. s0 is 1 units away. s1 is 0 units away. s2 is 1 units away. s3 in 2 units away. 
         
<Environment Details>: c0 contains 1 out of 3 onions. c0 is off. soup in c0 is not cooking. c1 contains 0 out of 3 onions. c1 is off. soup in c1 is not cooking.

Available Actions: [place onion in c0, place onion in c1., place onion on s0., place onion on s1., place onion on s2, place onion on s3., place onion on k12., wait., move away.]
\end{lstlisting}
\end{minipage}
    % \caption{Caption}
    \label{fig:enter-label}
\end{figure}

\newpage 
\clearpage

\section{Hanabi Implementation Details}
Hanabi is a card game in which all players are on the same team. The deck is made up of cards numbered 1 through 5, further divided into five different colors. Players are working together to create these numbered sequences (1 through 5) for each of the five colors. Each card played provides 1 point, and the goal is to obtain all 25 points (5 colors with 5 cards each) by completing the sequence for each color. The three actions a player can take include playing a card, discarding a card, and giving a hint (reveal) to a teammate. The challenge is that players cannot see their own cards. They may, however, see the cards that are in the hands of their teammate(s). This is where hints (reveals) come into play. Players are able to figure out which cards should be played through the reveal mechanism. You can hint (reveal) a teammate about a color or number in their hand, and this will point out to them all cards in their hand that have this color or number. The team starts with 8 reveal tokens but can recover used tokens when discarding a card or upon the completion of a stack. Playing a card carries a risk because the team loses completely if three incorrect cards are played. Further, an incorrectly played card gets sent to the discard pile. Each number of a particular color only has so many copies, so if, for example, the last copy of a green 3 is lost, then the team loses out on not only playing the green 3 but also the green 4 and 5. This is the risk of playing or discarding. With these risks, and since reveal tokens are scarce, players ideally try to make assumptions about implicit information that a reveal may offer, and, ideally, the team converges to particular conventions. Implicit communication, collaboration, and memory are key.  
\label{sec:hanabi-appendix}
\newpage
\subsection{Game Description}
We structure the game description of Hanabi into the overall objective, and the rules of the game. 
\begin{figure}[H]
    \centering
    \noindent 
\begin{minipage}{\columnwidth}
\begin{lstlisting}
The card game Hanabi has the following rules:
- The game uses a 50-card deck, divided into five colours (red (R), green (G), blue (B), yellow (Y), white (W)). Each color has cards of ranks 1 to 5. Each color has with three 1's, two 2's, two 3's, two 4's, one 5.
- Players have to create stacks of each color. Each color stack starts with a Rank 1 card and goes up one by one in ascending order up to Rank 5.  (e.g. Red Stack should go from R1 -> R2 -> R3 -> R4 -> R5). A card can only be played if it is the next in the incremental sequence for its color stack.
- Players can only see the other's hand, not their own.
- Players have plausible knowledge of their cards based on previously provided hints by the other player
- They can either play a card, give a reveal, or discard a card.
- Players can only chose an action from the Available Legal Actions.
***Actions:***
1. Reveal (Clue): Spend a reveal token to reveal cards with a particular color or rank. Revealing a color reveals all cards of that color in partner's hand. Revealing a rank reveals all cards with that rank in partner's hand. The game starts with 8 reveal tokens. If no token left, no more reveals can be given.
2. Discard: Discard a card to regain a reveal token and draw a new card. 
3. Play a Card: If a card played follows sequence in its color stack, it succeeds. Success of rank 5 card in any stack gives an additional reveal token. Failure discards the card, and loses a life. Playing a card you are unsure about is risky as it costs a life and you have only 3 lives. Before playing a card make sure that it's the next card in the sequence for that stack.
***The game ends when:***
- All five stacks are completed. 25 Points. 
- Three lives have been lost. 0 Points no matter how many cards have been placed in the stack. 

- After the last card from the deck is drawn and each player has had a final turn. Sum total of the top card ranks of each color stack.
I am Alice, playing the card game Hanabi with my partner Bob. 

At each time step I will provide you with the relevant information of the game. I will also provide you with the legal action, help me select the best next action. Remember I am playing as Alice. Format your response as Explanation: <brief explanation for selecting the move>\nAction:<selected move>. Do not say anything else. Got it?
\end{lstlisting}
\end{minipage}
\vspace{-.5em}
    % \caption{CAPTION}
    \label{fig:appendix-figure2}
\end{figure}

% \end{lstlisting}
% \end{minipage}
% \vspace{-.5em}
%     % \caption{CAPTION}
%     \label{fig:appendix-figure2}
% \end{figure}

% \begin{figure}[H]
%     \centering
%     \noindent 
% \begin{minipage}{\columnwidth}
% \begin{lstlisting}
% We have agreed to follow these conventions: Conventions:
% 1. **Card Layout:**
% - Cards are added to the right; the oldest card is on the left.
% - Positions are referenced from left to right.
% 2. **Clues:**
% - Two types of clues: Play Clue (play the card) and Save Clue (save for later).
% - If a Play Clue or Save Clue can't be given, players must discard.
% 3. **Play Clue:**
% - A play clue is revealing a card or cards in partners hand that are immediately playable on the stack by indicating their rank or color.
% 4. **Save Clue**
% - A save clue is used to save rank 5 cards, unique rank 2 cards and critical cards (only one of the kind left) 
% 5. **Do Not Repeat Known Information**
% - If a player already knows the color of their card, do not repeat the color in a clue. If a player already knows the rank of their card, do not repeat the rank in a clue.
% 5. **Prioritize Play Clues over Save Clues:**
% - Prefer giving Play Clues if both are viable options.
% 6. **Discard Without Fear:**
% - Discard confidently, as saving important cards is a team responsibility.
% 7. **Play with Fear:**
% - You can take risks and play a card even though you are not completely sure when you have 2 or 3 lives left. However when you have only 1 life left you should play a card only when you are sure that is goes next on the stack. 
\newpage
\subsection{State Description}
The state description includes the current Stack $S$, the player's knowledge of their cards $K$ (updated based on clues), the partner agent's cards $C$, the partner agent's knowledge of their cards $K'$ (updated based on previous clues), each card in the discard pile $d_i$, the remaining Life Tokens $l$, and reveal tokens $r$ and the remaining Deck Size $D$. 
We also precalculate the next card that goes on each stack since LLMs frequently fail to count which card should go next on each stack. 

\begin{figure}[H]
    \centering
    \noindent 
\begin{minipage}{\columnwidth}
\begin{lstlisting}
It is currently My (Alice) turn. 
Current Stacks: 
Red - Red 5, Yellow - Yellow 4, Green - Green 1, White - White 1, Blue - Blue 3 
My cards based on my knowledge:  
Card 0 could be: [Red, Yellow, Green, Blue] [1, 2, 3]
Card 1 could be: [Yellow, White, Blue] [1, 2, 3]
Card 2 could be: [Red] [2]
Card 3 could be: [Yellow, White, Blue] [1]
Card 4 could be: [Yellow, White, Blue] [1]
I can see Bob's Cards are:
[Card 0: Green 1]  
[Card 1: Green 2]
[Card 2: Green 4]
[Card 3: White 4]
[Card 4: Yellow 1]
Bob's Knowledge about his cards:  
Bob believes his Card 0 could be: [Yellow, Green, White, Blue] [1, 2, 4]
Bob believes his Card 1 could be: [Green, White] [1, 2, 4]
Bob believes his Card 2 could be: [Yellow, Green] [1, 2, 3, 4]
Bob believes his Card 3 could be: [Yellow, Green, White] [1, 2, 3, 4]
Bob believes his Card 4 could be: [Yellow, Green] [1, 2, 4]
Remaining Reveal Tokens: 1 
Remaining Lives: 1  
Deck Size: 3 
The discard pile is: [Red 4, Red 3, Red 1, Red 1, Yellow 5, Yellow 2, Yellow 4, Green 3, Green 2, Green 4, Green 3, Green 1, Green 5, Blue 5, Blue 3, Blue 4, Blue 4, Blue 1, White 4, White 3, White 2, White 5, White 3]
My Action History: [Discard Card 4, Play Card 0, Reveal Bob's Rank 3 Cards, Discard Card 0, Play Card 4]
The next playable cards for each stack are:  
Red Stack is Full.
Only Yellow 5 can be played on Yellow Stack  
Only Green 2 can be played on Green Stack  
Only White 2 can be played on White Stack  
Only Blue 4 can be played on Blue Stack

Available Actions: 
A. Reveal Bob's Yellow color cards
B. Reveal Bob's Green color cards
C. Reveal Bob's White color cards
D. Reveal Bob's rank 1 cards
E. Reveal Bob's rank 2 cards
F. Reveal Bob's rank 4 cards
G. Play my Card 0
H. Play my Card 1
I. Play my Card 2
J. Play my Card 3
K. Play my Card 4
L. Discard my Card 0
M. Discard my Card 1
N. Discard my Card 2
O. Discard my Card 3
P. Discard my Card 4
\end{lstlisting}
\end{minipage}
\vspace{-.5em}
    % \caption{CAPTION}
    \label{fig:appendix-figure2}
\end{figure}

\newpage 
\clearpage
\section{Examples of prompts of LLMs used in the Agent framework}

\subsection{Hanabi LLM Prompt}

\begin{figure}[H]
    \centering
\noindent \begin{minipage}{\columnwidth}
\begin{lstlisting}
The card game Hanabi has the following rules:
    {self.rules}
I am {self.player_names[self.player_id]}, playing the card
game Hanabi with {self.player_names[1 - self.player_id]}. 
At each time step I will provide you with the relevant 
information of the game. I will also provide you with the 
legal action, help me select the best next action.
Remember I am playing as {self.player_names[self.player_id]}. 
Format your response as 
Explanation: <brief explanation for selecting the move>
Action:<selected move>. 
Do not say anything else. Got it?
\end{lstlisting}
\end{minipage}
    % \caption{Caption}
    \label{fig:enter-label}
\end{figure}

\subsection{Prompt for Theory of Mind Reasoning step}

\begin{figure}[H]
    \centering
\noindent \begin{minipage}{\columnwidth}
\begin{lstlisting}
The card game Hanabi has the following rules:
{self.rules}
I am {self.player_names[self.player_id]}, playing the card game Hanabi with {self.player_names[1-self.player_id]}. 
You are a Theory of Mind inference agent for our game. You will be provided with my partner's selected action and my latest state information after my partner took their action. You will provide me with two things: 1.  An explanation for my partner's previous action along with their intention and implicit communication. 2. What is the best information for me to give my partner based on their knowledge? 
Format your response as:
Partner Action Explanation:<1 sentence explanation of partner action>
Clue Suggestion:<What information (specify rank or color) should I reveal to my partner based on their knowledge>.
\end{lstlisting}
\end{minipage}
    % \caption{Caption}
    \label{fig:enter-label}
\end{figure}

\subsection{Prompt for Answer Verification Step} 

\begin{figure}[H]
    \centering
\noindent \begin{minipage}{\columnwidth}
\begin{lstlisting}
You are an action verification agent for games. I will provide 
you with an action and you need to check whether the action 
satisfies the criteria: 
1. Rule Following: It follows to the rules of the game. 
2. Safety: It won't lead to the game ending immediately.
Think about the action, the current state of the stack and the 
available lives and reveal tokens. 
End you response with "Verification: Okay" if selected action
follows ***both*** criteria and "Verification: Not Okay" 
otherwise. Restrict your response to 4-5 sentences.
\end{lstlisting}
\end{minipage}
    % \caption{Caption}
    \label{fig:enter-label}
\end{figure}

\newpage 
\subsection{Ablation on Theory of Mind Reasoning and Verification Steps}
\begin{table}[h]
\centering
\small
\begin{tabular}{llc}
\toprule
\textbf{Method} & \textbf{Score} & \textbf{Bomb Rate} \\
\midrule
ToM + GPT-4-turbo + V  & \(\mathbf{13.33} \pm 0.88\) & $0.00$ \\
GPT-4-turbo + V & \(10.33 \pm 0.88\)  & $0.00$\\
GPT-4-turbo & \(4.33 \pm 0.67\) & $1.00$ \\
\bottomrule
\end{tabular}
\caption{Ablation study of LLM agents on Hanabi Challenge (w. GPT-4-turbo). Answer Verification (AV) markedly enhances overall performance by ensuring that actions that make incorrect assumptions are filtered out. The explicit Theory of Mind (ToM) reasoning provides further improvements by directly interpreting partner clues and requirements. }
\label{tab:hanabi_ablations}
\end{table}

% \FloatBarrier

% \begin{table*}[!t]
% \centering
% \small
% \begin{tabular}{lcccc}
% \toprule
% & \multicolumn{2}{c}{\textbf{Collab Escape}} & 
% \multicolumn{2}{c}{\textbf{Collab Capture}} \\
% \cmidrule(lr){2-3} \cmidrule(lr){4-5}

% \textbf{Agent} & Capture Rate & Avg. Turns & Escape Rate & Avg. Turns \\ 

% \midrule
% GPT-4-turbo    & 0.83 & 4.60 & 1.00 & 3.5 \\
% GPT-4-turbo \textbf{(w/out PI)}   & 0.50 & 2.00 & 1.00 & 4.75 \\
% GPT-4o    & 0.67 & 4.00 & 1.00 & 7.17 \\
% %GPT-4o-mini    & 1.00 & 7..33 & 0.67 & 25 \\
% % -- Aux. Modules  & -    & -     &  1.00   &  24.67    \\
% GPT-3.5-turbo  & 0.33 & 2.50 & 0.67 & 8.38  \\
% Mixtral-8x7b   & 0.50 & 7.67 & 0.92 & 7.55 \\

% \midrule
% Greedy Baseline& 0.00 & N.A. &  0.50 & 6.00 \\
% \bottomrule
% \end{tabular}
% \caption{Comparison of different LLMs on CollabCapture and CollabEscape with the ReAct \cite{yao2023reactsynergizingreasoningacting} agent framework. An ablation of the partner-inference (PI) module of the GPT-4-based agent was performed. The PI module requires the agent to explicitly reflect on their partner's intentions before thinking what move is best for themselves. Agents with this explicit PI module perform more consistently on both environments.}

% \label{table:model_combined}
% \end{table*}

% \FloatBarrier

\clearpage

\newpage
\section{Generating Questions for CoordinationQA}
\label{app:gqc}
\subsection{Environment Comprehension Questions}
\label{app:ec}
The Environment Comprehension (EC) questions are indirect formulations regarding spatial aspects of the layout. In order for an agent to correctly answer an EC question, they must have an understanding of the dynamic details of the current state and the rules of the game and exhibit spatial awareness. As such, when creating the EC questions, we carefully comb through a given scenario in search of salient points to probe an agent's understanding of the given environment. Some examples include:

\noindent \begin{minipage}{\columnwidth}
\begin{lstlisting}
<Inventory>: I am holding nothing. Bob is holding onion.
<My location information:> o0 is 1 units away. o1 is 0 units away. p0 is 1 units away. d0 is inaccessible. c0 is inaccessible. c1 is inaccessible. s0 is 1 units away. s1 is 0 units away. s2 is 1 units away. 
<Bob's location information>: o0 is inaccessible. o1 is inaccessible. p0 is inaccessible. d0 is 2 units away. c0 is 0 units away. c1 is 0 units away. s0 is 0 units away. s1 is 1 units away. s2 is 2 units away. 
<Environment Details>: c0 contains 3 out of 3 onions. c0 is on. soup in c0 is still cooking. c1 contains 0 out of 3 onions. c1 is off. soup in c1 is not cooking. s0 is empty. s1 contains onion. s2 is empty. Closest empty kitchen counter k1 is 1 units away. 

How many onions are still needed to fill up c0? 
Available Answers: 
A. 4 or more
B. 3
C. 2
D. 1
E. 0
\end{lstlisting}
\end{minipage}

\noindent\begin{minipage}{\columnwidth}
\begin{lstlisting}
My name is Alice. I am in room 1. Bob is in room 6. I was fixing the generator and there is only one more fix needed, which could be done before getting caught. Currently, we have information that the killer will move to the room 1 after this turn. Generator in room 1 still needs 1 fix. Generator in room 2 is fixed. The exit gate is closed. 

If I fix generator 1, is Bob in a position to escape? 

Available Answers: 
A. Yes, he's only one room away from the gate when it opens.
B. No, the killer is blocking his path to the exit gate.
C. No, we stil need to fix generator 2.
\end{lstlisting}
\end{minipage}
\newpage

\subsection{Theory of Mind Reasoning Questions}
\label{app:tom}
There are two primary question types in Hanabi for ToM Reasoning questions. In the first type, we ask the LLM about what information the partner agent needs, while in the second type, we ask it to make inferences about the partner agent's last action. For all games apart from Hanabi, the ToM questions ask the models to predict the next intended action of the partner agent

\subsubsection{Hanabi Question Type-1}

\noindent \begin{minipage}{\columnwidth}
\begin{lstlisting}
It is currently My (Alice) turn. Current Stacks: Red - Red 0, Yellow - Yellow 0, Green - Green 0, White - White 0, Blue - Blue 0
My cards based on my knowledge:  
Card 0 could be: [Red, Yellow, Green, White, Blue] [1, 2, 3, 4, 5]
Card 1 could be: [Red, Yellow, Green, White, Blue] [1, 2, 3, 4, 5]
Card 2 could be: [Red, Yellow, Green, White, Blue] [1, 2, 3, 4, 5]
Card 3 could be: [Red, Yellow, Green, White, Blue] [1, 2, 3, 4, 5]
Card 4 could be: [Red, Yellow, Green, White, Blue] [1, 2, 3, 4, 5]
I can see Bob's Cards are:  
[Card 0: Red 3]  
[Card 1: White 1]  
[Card 2: Green 3]  
[Card 3: White 4]  
[Card 4: Blue 4]  
Bob's Knowledge about his cards:  
Bob believes his Card 0 could be: [Red, Yellow, Green, White, Blue] [1, 2, 3, 4, 5]
Bob believes his Card 1 could be: [Red, Yellow, Green, White, Blue] [1, 2, 3, 4, 5]
Bob believes his Card 2 could be: [Red, Yellow, Green, White, Blue] [1, 2, 3, 4, 5]
Bob believes his Card 3 could be: [Red, Yellow, Green, White, Blue] [1, 2, 3, 4, 5]
Bob believes his Card 4 could be: [Red, Yellow, Green, White, Blue] [1, 2, 3, 4, 5]
Remaining Reveal Tokens: 8 
Remaining Lives: 3  
Deck Size: 40 
The discard pile is: []
My Action History: []
The next playable cards for each stack are:  
Only Red 1 can be played on Red Stack  
Only Yellow 1 can be played on Yellow Stack  
Only Green 1 can be played on Green Stack  
Only White 1 can be played on White Stack  
Only Blue 1 can be played on Blue Stack
\end{lstlisting}
\end{minipage}
\newpage

\noindent \begin{minipage}{\columnwidth}
\begin{lstlisting}
What information about his cards should I reveal to my partner so that he knows to play a card on his turn?
Available Answers:
A. Reveal Bob's Red color cards.
B. Reveal Bob's White color cards.
C. Reveal Bob's Green color cards.
D. Reveal Bob's Blue color cards.
E. Reveal Bob's rank 1 cards.
F. Reveal Bob's rank 3 cards.
G. Reveal Bob's rank 4 cards.
\end{lstlisting}
\end{minipage}

\subsubsection{Hanabi Question Type-2}
\noindent \begin{minipage}{\columnwidth}
\begin{lstlisting}
It is currently My (Alice) turn. Current Stacks: Red - Red 1, Yellow - Yellow 2, Green - Green 1, White - White 4, Blue - Blue 3 
My cards based on my knowledge:  
Card 0 could be: [Red, Yellow, Green, White, Blue] [1, 2, 3, 4, 5]
Card 1 could be: [Red, Yellow, Green, White, Blue] [1, 2, 3, 5]
Card 2 could be: [Red, Yellow, Green, White, Blue] [1, 2, 3, 5]
Card 3 could be: [Red, Yellow, Green, Blue] [3]
Card 4 could be: [White] [5]
I can see Bob's Cards are:  
[Card 0: Yellow 1]  
[Card 1: Blue 1]  
[Card 2: Blue 1]  
[Card 3: Red 3]  
[Card 4: Green 3]  
Bob's Knowledge about his cards:  
Bob believes his Card 0 could be: [Red, Yellow, Green, White, Blue] [1, 2, 3, 4, 5]
Bob believes his Card 1 could be: [Red, Yellow, Green, White, Blue] [1, 2, 3, 4, 5]
Bob believes his Card 2 could be: [Red, Yellow, Green, White, Blue] [1, 2, 3, 4, 5]
Bob believes his Card 4 could be: [Red, Yellow, Green, Blue] [3]
Remaining Reveal Tokens: 1 
Remaining Lives: 2  
Deck Size: 25 
The discard pile is: [Yellow 4, Blue 2, Blue 3, White 2, White 3, White 4]
My Action History: [Reveal Bob's Rank 2 Cards, Reveal Bob's Rank 5 Cards, Reveal Bob's Rank 2 Cards, Play Card 1, Reveal Bob's Rank 1 Cards,  Discard Card 0, Reveal Bob's Rank 3, Reveal Bob's Rank 2, Reveal Bob's Rank 2 Cards, Reveal Bob's Rank 1 Cards, Discard Card 3, Reveal Bob's White Color Cards, Discard Card 1]
The next playable cards for each stack are:  
Only Red 2 can be played on Red Stack  
Only Yellow 3 can be played on Yellow Stack  
Only Green 2 can be played on Green Stack  
Only White 5 can be played on White Stack  
Only Blue 4 can be played on Blue Stack


What can I infer from my partner's previous action? 
Available Answers:
A. I should Play Card 0
B. I should Play Card 1
C. I should Play Card 2
D. I should Play Card 3
E. I should Play Card 4
F. I should Discard Card 0
G. I should Discard Card 1
H. I should Discard Card 2
I. I should Discard Card 3
J. I should Discard Card 4
\end{lstlisting}
\end{minipage}
\newpage

\subsubsection{Other Games}

\noindent \begin{minipage}{\columnwidth}
\begin{lstlisting}
<Inventory>: I am holding onion. Bob is holding nothing. 

<My Location Information>: o0 is 0 units away. o1 is 1 units away. p0 is 3 units away. c0 is 6 units away blocked by Bob. c1 is 7 units away. d0 is 4 units away. s0 is 1 units away. s1 is 0 units away. s2 is 1 units away. s3 in 2 units away. Closest empty kitchen counter k12 is 1 units away. 

<Bob's Location Information>: o0 is blocked by Alice. o1 is 7 units away. p0 is 3 units away. c0 is 0 units away. c1 is 1 units away. d0 is 4 units away. s0 is 1 units away. s1 is 0 units away. s2 is 1 units away. s3 in 2 units away. 
         
<Environment Details>: c0 contains 1 out of 3 onions. c0 is off. soup in c0 is not cooking. c1 contains 0 out of 3 onions. c1 is off. soup in c1 is not cooking. 

         
What action does my partner intend to take? 
Available Actions: 
A. pick up onion from o0.
B. pick up onion from o1.
C. pick up plate from p0.
D. pick up onion from s0.
E. pick up onion from s1.
F. pick up onion from s2.
G. pick up onion from s3.
H. pick up plate from s0.
I. pick up plate from s1.
J. pick up plate from s2.
K. pick up plate from s3.
L. wait.
M. move away.
\end{lstlisting}
\end{minipage}

\subsection{Joint Planning Questions}
\label{app:jp}
Joint planning questions are effectively the same questions that the LLM solves when they are part of an agentic framework. For each scenario, we ask the LLM to answer the question: "What is the best next action?".

\noindent \begin{minipage}{\columnwidth}
\begin{lstlisting}
I (Alice) am in Room 6. Bob is in Room 1. Thief is in Room 2. 
Door between Room 1 and 2 is closed. Door between Room 3 and 4 is closed. 

What action should I take next? 
Available Actions: 
A. Move to Room 1
B. Move to Room 5
C. Move to Room 9
D. Stay in current Room
\end{lstlisting}
\end{minipage}

\newpage 

\section{CollabCapture and CollabEscape}
\label{app:collab}

\subsection{CollabCapture}
\label{app:collabCapture}
\begin{figure*}[hbp]
    \centering
    \includegraphics[width=\textwidth]{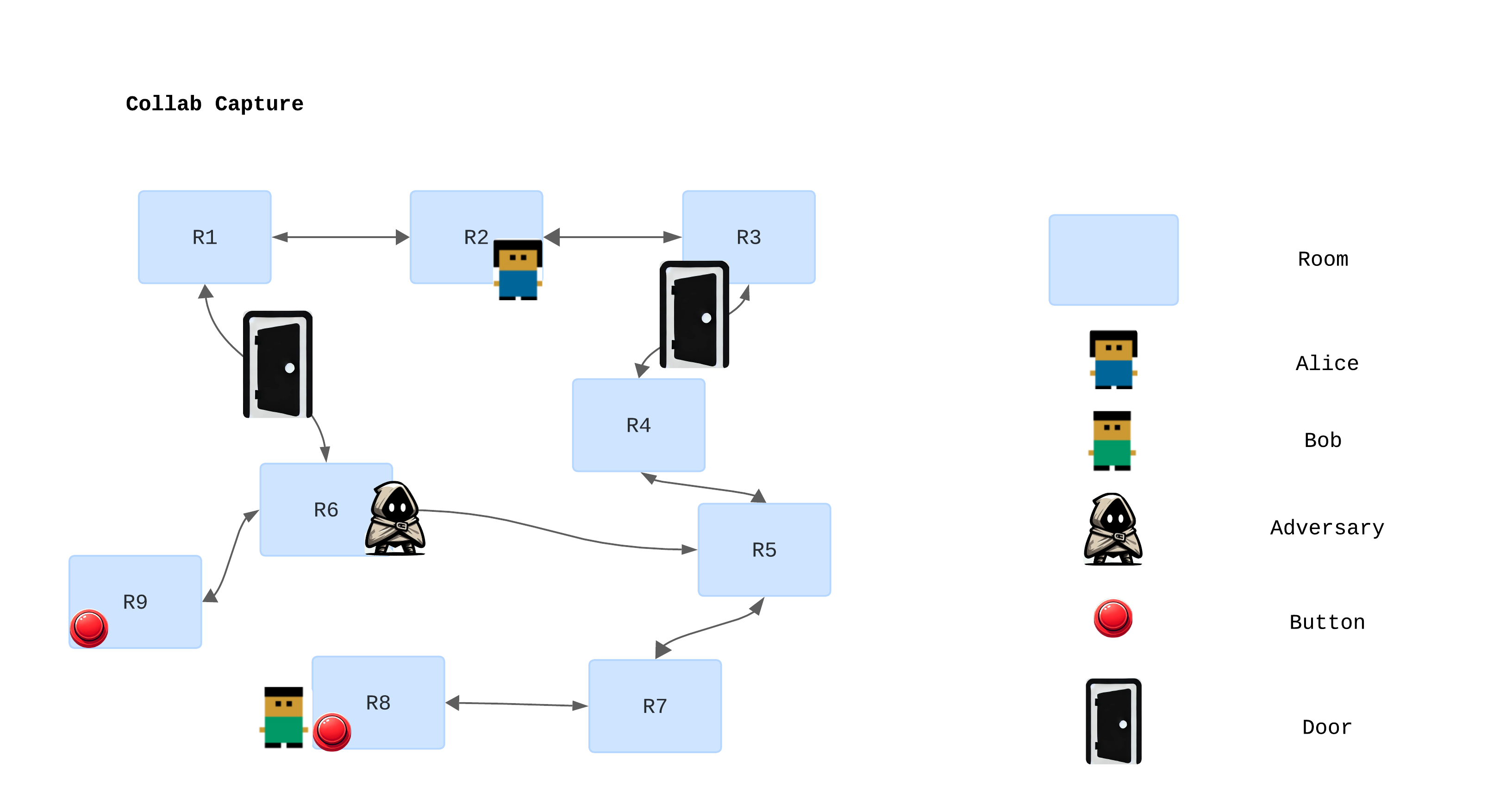}
    \caption{Map layout for CollabCapture}
    \label{fig:collab_map}
\end{figure*}

CollabCapture places two agents in a set of interconnected rooms and doors. Their goal is to capture an adversary within the smallest number of moves possible. The adversary moves away from agents in a greedy fashion, but the agents have the ability to close and open doors while the adversary does not. The doors are controlled by a corresponding button that is in another location on the map. (See Figure \ref{fig:collab_map})

CollabCapture contains one layout with four scenarios created by controlling the gate states (open/closed). This corresponds to cases where players need to: 1. Pincer their opponent through coordination 2. One agent needs to enable the other agent by choosing not to chase the agent but rather open the door to allow the other agent to capture the adversary. 3. One agent needs to disable the adversary by closing a door, allowing the other agent to catch them. CollabCapture is based on the classic task of Pursuit Evasion from the perspective of the pursuers. This is representative of a common-payoff task as the only objective is capturing the adversary with no mixed incentives (Akin to deliveries in Overcooked, where all chefs get a common payoff with no preference for the one making the delivery).

\subsection{CollabEscape}
\label{app:collabEscape}
\begin{figure*}[hbp]
    \centering
    \includegraphics[width=\textwidth]{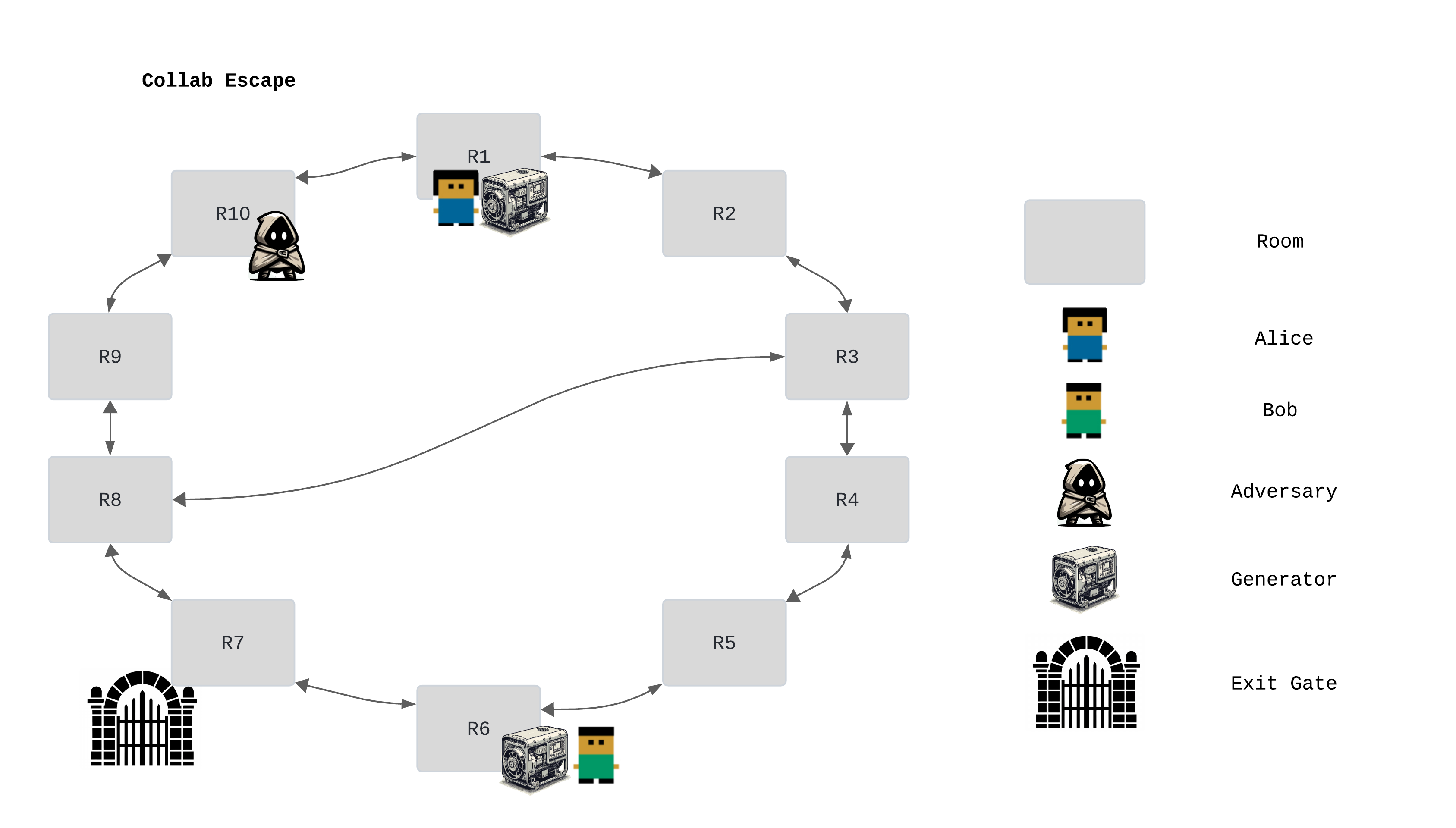}
    \caption{Map layout for CollabEscape}
    \label{fig:collab_esc_map}
\end{figure*}

CollabEscape also places two agents in a set of interconnected rooms and doors. Their goal in this environment, however, is to escape an adversary that is looking to catch them. The map has two generators that power an exit gate, both of which need to be fixed. Upon fixing both generators, the players must then escape by reaching the exit gate. Only one player needs to reach the exit gate in order for them both to win. If the adversary catches either of them, however, they both lose. (See Figure \ref{fig:collab_esc_map})

In CollabEscape, we have one layout with two scenarios created by varying the starting positions of the two agents. Depending on their proximity to the adversary/generators, players need to apply strategies such as luring the adversary away from the partner, choosing to continue fixing the generators while sacrificing for the partner’s safety, and manipulating the movement of the adversary. This is also representative of a common payoff task, as players are commonly rewarded if any one of them escapes, introducing roles of explicit assistance and sacrifice for a higher common payoff.

\end{document}